\documentclass{article}

\usepackage{microtype}
\usepackage{graphicx}
\usepackage{subfigure}
\usepackage{booktabs} % for professional tables

\usepackage{hyperref}

\usepackage{adjustbox}

\usepackage{pifont}
\usepackage{multirow}
\usepackage{amsmath}

\newcommand{\cmark}{\ding{51}}
\newcommand{\xmark}{\ding{55}}

\DeclareMathOperator*{\vleft}{<Vleft>}
\DeclareMathOperator*{\vright}{<Vright>}

\usepackage[accepted]{icml2020}

\icmltitlerunning{Imitation Learning Approach for AI Driving Olympics Trained on Real-world and Simulation Data Simultaneously}

\begin{document}

\twocolumn[
\icmltitle{Imitation Learning Approach for AI Driving Olympics \\ Trained on Real-world and Simulation Data Simultaneously}

% It is OKAY to include author information, even for blind
% submissions: the style file will automatically remove it for you
% unless you've provided the [accepted] option to the icml2020
% package.

% List of affiliations: The first argument should be a (short)
% identifier you will use later to specify author affiliations
% Academic affiliations should list Department, University, City, Region, Country
% Industry affiliations should list Company, City, Region, Country

% You can specify symbols, otherwise they are numbered in order.
% Ideally, you should not use this facility. Affiliations will be numbered
% in order of appearance and this is the preferred way.
\icmlsetsymbol{equal}{*}

\begin{icmlauthorlist}
\icmlauthor{Mikita Sazanovich}{jbr,hse}
\icmlauthor{Konstantin Chaika}{jbr,leti}
\icmlauthor{Kirill Krinkin}{jbr,leti}
\icmlauthor{Aleksei Shpilman}{jbr,hse}
\end{icmlauthorlist}

\icmlaffiliation{jbr}{JetBrains Research, St Petersburg, Russia}
\icmlaffiliation{hse}{National Research University Higher School of Economics, St Petersburg, Russia}
\icmlaffiliation{leti}{SPbETU "LETI", St Petersburg, Russia}

\icmlcorrespondingauthor{Mikita Sazanovich}{mikita.sazanovich@jetbrains.com}

% You may provide any keywords that you
% find helpful for describing your paper; these are used to populate
% the "keywords" metadata in the PDF but will not be shown in the document
\icmlkeywords{Machine Learning, ICML}

\vskip 0.3in]

% this must go after the closing bracket ] following \twocolumn[ ...

% This command actually creates the footnote in the first column
% listing the affiliations and the copyright notice.
% The command takes one argument, which is text to display at the start of the footnote.
% The \icmlEqualContribution command is standard text for equal contribution.
% Remove it (just {}) if you do not need this facility.

\printAffiliationsAndNotice{}  % leave blank if no need to mention equal contribution
% \printAffiliationsAndNotice{\icmlEqualContribution} % otherwise use the standard text.

%%%%%%%%%%%%%%%%%%%%%%%%%%%%%%%%%%%%%%%%%%%%%%%%%%%%%%%%%%%%%%%%%%%%%%%%%%%%%%%%
\begin{abstract}
In this paper, we describe our winning approach to solving the Lane Following Challenge at the AI Driving Olympics Competition through imitation learning on a mixed set of simulation and real-world data. AI Driving Olympics is a two-stage competition: at stage one, algorithms compete in a simulated environment with the best ones advancing to a real-world final. One of the main problems that participants encounter during the competition is that algorithms trained for the best performance in simulated environments do not hold up in a real-world environment and vice versa. Classic control algorithms also do not translate well between tasks since most of them have to be tuned to specific driving conditions such as lighting, road type, camera position, etc. To overcome this problem, we employed the imitation learning algorithm and trained it on a dataset collected from sources both from simulation and real-world, forcing our model to perform equally well in all environments. 
% This approach has yielded us the first place at the AI Driving Olympics 2019. In the paper, we both describe the approach and present experimental results that show its superior performance.
\end{abstract}

%%%%%%%%%%%%%%%%%%%%%%%%%%%%%%%%%%%%%%%%%%%%%%%%%%%%%%%%%%%%%%%%%%%%%%%%%%%%%%%%
\section{Introduction}

\begin{figure}
    \centering
    \includegraphics[width=\linewidth]{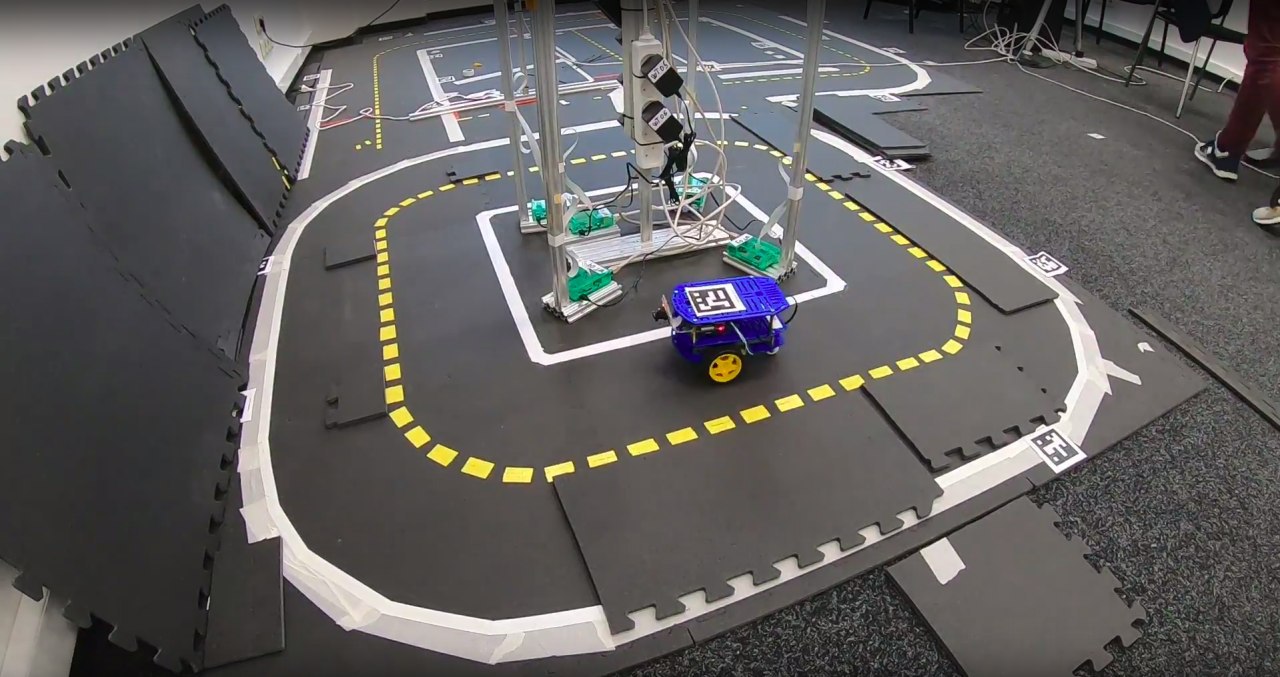}
    \caption{The Duckietown vehicle and our training environments: the simple loop (foreground) and roads with intersections (background).}
    \label{fig:setup}
\end{figure}

The AI Driving Olympics (AIDO) \cite{zilly2019aido} aims to help with developing algorithms for self-driving cars and is built on the Duckietown platform \cite{paull17duckietown}. Since development on full-sized autonomous cars is costly, many researchers resort to developing and testing their algorithms on miniaturized versions of streets and cities. In Duckietown AI Driving Olympics Lane Following Challenge, the goal is to control just such a miniature car on a road in a miniature city environment (Figure~\ref{fig:setup}).

The car receives images from a single camera mounted in front, and the algorithm needs to output voltages for each of the two front-wheel drives. Algorithms are judged by how far the car can go without driving off the road.

% The competition is divided into two stages:

% \begin{enumerate}
% \item Simulator stage. Organizers provide a simulated environment, for the ease of training and testing of the algorithms. Top 16 performing teams are then advance to the second, real-world, stage.
% \item Real-world stage. Algorithms from top 16 teams of the first stage compete in a real-world environment. It is important to note, that the same algorithms from the first stage compete in the real-world stage, and teams are not allowed to change their algorithm between stages. So the algorithm needs to perform well in both the simulated and the real-world environment.
% \end{enumerate}

The competition is divided into two stages: simulation and real-world. A single algorithm needs to perform well in both. It was quickly identified, that one of the major problems is the simulation to real-world transfer. Many algorithms trained in simulated environment performed very poorly in the real world, and many classic control algorithms that are known to perform well in a real-world environment, once tuned to that environment, do not perform well in the simulation. Simulation to real-world transfer problem is well-known, including in the area of self-driving vehicles \cite{pan2017virtual, csurka2017survey}. Some approaches suggests randomizing the domain for the transfer \cite{tobin2017domain}.

% Simulation to real-world transfer problem is well-known, including in the area of self-driving vehicles \cite{pan2017virtual, tobin2017domain, csurka2017survey}.

% \begin{figure*}
%     \centering
%     \includegraphics[width=\linewidth]{logging.jpg}
%     \caption{View from the robot's camera and logged voltages of the left and the right wheel.}
%     \label{fig:logging}
% \end{figure*}

In this paper, we propose a novel method of training a neural network model that can perform well in diverse environments, such as simulations and real-world environment. To that end, we have trained it through imitation learning on a dataset compiled from four different sources:

\begin{enumerate}
\item Real-world dataset provided by the Duckietown \mbox{organizers}.
\item Simulation dataset on a simple loop map where the car was driven by a tuned PD controller.
\item Simulation dataset on an intersection map, the car was driven by a tuned PD controller. 
\item Real-world dataset collected by us in our environment, where the car was driven by a tuned PD controller.
\end{enumerate}

In the general sense, our network have learned to imitate the behaviour of both human experts and automatic controllers that are tuned to a specific task, and training it on data from all aforementioned sources have helped us achieve the first place at AIDO Lane Following Challenge 2019.

\section{Dataset Generation}
\label{sec:dataset-generation}

Our data consists of images from a single camera mounted in front of the car. For every image, the model needs to predict an expert's action, be that human driver or a tuned algorithm. The action is coded as a tuple of two real-valued numbers, $\vleft,\vright$, both in $[-1,1]$, that determine the drive voltages of the left and the right wheely. 

\subsection{Simulation data}
To extract data in the image$\rightarrow$voltage format we just had to add the image and voltage saving module to the simulator provided by the Duckietown team. Using the position relative to the lane, we can use a tried and true proportional-derivative (PD) controller \cite{nise2007control}. We have, however, needed to fine-tune the controller's parameters. 

After fine-tuning to the best lane following performance, we have collected images and voltages from simulation runs. We have done it in two maps, one is just a simple loop so that our model learns to follow the lane in simple conditions as best as it can, another is a more complex map with intersections and other complications that aims to improve model performance in similar complex surroundings. We label these two datasets as SIM-LP and SIM-IS respectively.

\subsection{Real-world data provided by Duckietown organizers}

Some data for imitation learning is provided by the organizers of the competition \cite{paull17duckietown}. These logs are available at logs.duckietown.org. These logs may be uploaded by any member of the Duckietown community and vary in environment conditions, such as map configuration, lighting, etc. We label this dataset as REAL-DT.

\subsection{Real-world data recorded in-house}

Second real-world data source is in-house recordings in our own Duckietown environment in various scenarios. We label this dataset as REAL-IH.

% \subsubsection{Recording scenarios}

We aimed to collect data with as many possible situations as we can. These situations should include twists in the road, driving in circles clockwise and counterclockwise, and so on. We have also tried to diversify external factors such as scene lighting, items in the room that can get into the camera's field of view, roadside objects, etc. If we keep these conditions constant, our model may overfit to them and perform poorly in a different environment. To fight that, we have recorded several tracks for the same road configuration but under different conditions. Road configurations mirrored that of the simulation runs so that our model can see similar situations and learn to extract the most relevant features, while ignoring the domain specific ones.
% For example, if simulation has areas of several left-hand and right-hand turns in a row, our real-world environment should have similar areas as well.

% \subsubsection{Robot control and data acquisition}

\begin{figure*}
    \centering
    \includegraphics[height=1.2in]{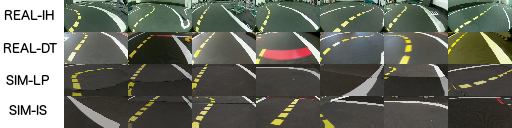}
    \caption{Preprocessed samples (from top to bottom) from real-world in-house (REAL-IH), real-world Duckietown (REAL-DT), simulation with one loop and no intersections (SIM-LP), and simulation with intersection (SIM-IS) datasets.}
    \label{fig:samples}
\end{figure*}

\begin{figure*}
    \centering
    \includegraphics[height=1.2in]{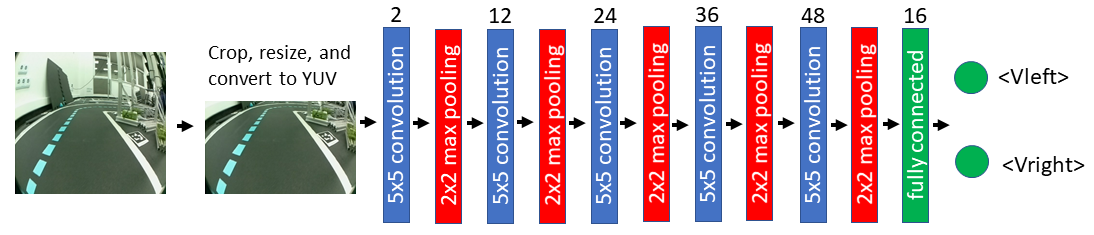}
    \caption{Architecture of our network. Numbers in the upper row denote numbers of convolutional maps and number of neurons for fully connected layer. $\vleft$ and $\vright$ denote the voltage output to the left and the right wheel.}
    \label{fig:network}
\end{figure*}

To control the robot in our environment, we have used the PD controller \cite{nise2007control} that gets information about the robot's position from the camera by detecting lanes by their color and then steers the robot, so it would keep to the center of the lane. The lane detection was calibrated for every lighting condition since different lighting changes the color scheme of the image input. One modification we have made to the standard PD algorithm is how it approaches intersections and sharp turns. Since most Duckietown turns and intersections are standard-shaped, we hard-coded the robot's motion in these situation. 

One of the key aspects of our solution is that we have not excluded imperfect trajectories, ones that would slightly go out of bounds of the lane for example. This would help our end model to function in some unexpected situations and perform actions needed to return to the lane and continue safe movement. These imperfections in robot's actions also increase the robustness of the model.

\section{Neural network architecture and training}

The real-world camera and the simulation model output 640x480 RGB images to be used by the control policy. As a preprocessing step, we remove the top third of the image, since it mostly contains the background and does not affect the control decisions, resize the image to 64x32 pixels and convert it into the YUV colorspace. Figure~\ref{fig:samples} displays a subset of preprocessed samples.

The model choices have been in part governed by the method proposed in \cite{bojarski2016nvidia}. In particular, we have followed the general CNN architecture the authors propose. We have used 5 convolutional layers with a small number of filters, followed by 2 fully-connected layers. The small size of the network is not only due to it being less prone to overfitting, but we were also required by the competition organizers to create a model that can run on a single CPU. The full model's architecture is shown in Figure~\ref{fig:network}.

We have also incorporated Independent-Component (IC) layers from the recent paper \cite{chen2019iclayers}. These layers aim to make the activations of each layer more independent by combining two popular techniques, BatchNorm \cite{ioffe2015batch} and Dropout \cite{Srivastava2014DropoutAS}. For convolutional layers, we substitute Dropout with Spatial Dropout \cite{Tompson2015EfficientOL} which has been shown to work better with them. The Spatial Dropout layer zeros out the whole map of neurons with some predefined probability. We set this probability to $0.01$ for the first convolutional layer and $0.05$ for the following four convolutional layers. For the first fully-connected layer, we use Dropout with the dropout rate of $0.05$. The model outputs two values for voltages of the left and the right wheel drives. We use the mean square error (MSE) as our training loss.

We have tried to experiment with sequential data, such as few consecutive input frames, but it has not improved the performance, so our final approach operates only on the current frame. 

As for the model training regime, we have trained it on the dataset where all data from all sources are combined in equal proportions. We have trained several models on the combined training dataset, consisting of 20294 images in total. We have computed MSE for all validation sets, consisting of 8698 images in total. We have selected the model with the lowest averaged across validation sets MSE.

\section{Results}

In this section, we first present the results of training different approaches on different data sources and then the results of applying the trained approaches to control a car in both simulated and real-world scenarios.

We consider the following approaches: the classic control algorithm provided by the Duckietown organizers (CC), the model trained on data from real-world sources only (REAL), the model trained on data from simulation sources only (SIM), the model trained on all data sources (HYBRID).

\subsection{Training evaluation}

For the training evaluation, we compute the mean square error (MSE) of the left and the right voltage outputs on the validation set of each data source. We rerun the training process for each method 5 times with different randomly initialized weights, and report averaged values. Table~\ref{table:offline-eval} shows the results (in order from top) for the model trained on all data sources, on real-world data sources only, on simulation data sources only, and  on each of the data sources separately. For more in-depth description of the datasets, refer to Section~\ref{sec:dataset-generation}. As we can see from Table~\ref{table:offline-eval}, while training on a single dataset sometimes achieves lower error on the same dataset than our hybrid approach, we do achieve the best performance in terms of the average MSE on all four datasets. We can also see that our method performs on par with the best single methods. In terms of the average error it outperforms the closest one tenfold. This demonstrates definitively the high dependence of MSE on the training method, and highlights the differences between the data sources.

\begin{table}[t!]
\begin{adjustbox}{width=\columnwidth,center}
\begin{small}
\begin{tabular}{|c|c c c c c|}
 \hline
 \multirow{2}{*}{Method} & \multirow{2}{*}{REAL-IH} & \multirow{2}{*}{REAL-DT} & \multirow{2}{*}{SIM-LP} & \multirow{2}{*}{SIM-IS} & \multirow{2}{*}{AVG} \\
& & & & & \\
 \hline
 HYBRID & 0.0178 & 0.0070 & 0.0108 & 0.0209 & \textbf{0.0141} \\
 \hline
 REAL & 0.0168 & 0.0064 & 0.3712 & 0.3564 & 0.1877 \\
 \hline
 SIM & 0.1325 & 0.4285 & \textbf{0.0097} & \textbf{0.0183} & 0.1473 \\
 \hline
 REAL-IH & \textbf{0.0167} & 0.0173 & 0.4196 & 0.4011 & 0.2137 \\
 \hline
 REAL-DT & 0.0379 & \textbf{0.0059} & 0.3682 & 0.3491 & 0.1903 \\
 \hline
 SIM-LP & 0.3428 & 0.4785 & 0.0105 & 0.0684 & 0.2250 \\
 \hline
 SIM-IS & 0.1197 & 0.4181 & 0.0178 & 0.0199 & 0.1439 \\
 \hline
\end{tabular}
\end{small}
\end{adjustbox}
\caption{MSE on validation set of each data source.}
\label{table:offline-eval}
\end{table}

\begin{table}[t!]
\begin{adjustbox}{width=\columnwidth,center}
\begin{small}
\begin{tabular}{|c|c c c c c c c|}
 \hline
 \multirow{2}{*}{Method} & \multicolumn{2}{c}{Scenario 1} & \multicolumn{2}{c}{Scenario 2} & \multicolumn{2}{c}{Scenario 3} & Total\\
 & Tiles & Time & Tiles & Time & Tiles & Time & Tiles \\
 \hline
 HYBRID & 3.10 & 15 & 3.32 & 15 & 3.29 & 15 & 9.71 \\
 \hline
 REAL & 2.92 & 15 & 2.76 & 15 & 3.16 & 15 & 8.84 \\
 \hline
 SIM & \textbf{3.53} & 15 & \textbf{3.97} & 15 & \textbf{4.01} & 15 & \textbf{11.51} \\
 \hline
 CC & 2.27 & 15 & 2.28 & 15 & 2.28 & 15 & 6.83 \\
 \hline
\end{tabular}
\end{small}
\end{adjustbox}
\caption{Simulation closed-loop validation performance.}
\label{table:simulation-val-eval}
\end{table}

\begin{table}[t!]
\begin{adjustbox}{width=\columnwidth,center}
\begin{small}
\begin{tabular}{|c|c c c c c c c|}
 \hline
 \multirow{2}{*}{Method} & \multicolumn{2}{c}{Scenario 1} & \multicolumn{2}{c}{Scenario 2} & \multicolumn{2}{c}{Scenario 3} & Total \\
 & Tiles & Time & Tiles & Time & Tiles & Time & Tiles \\
 \hline
 HYBRID & \textbf{6.88} & 37 & \textbf{8.99} & 46 & \textbf{3.41} & 21 & \textbf{19.28} \\
 \hline
 REAL & 6.65 & 33 & 8.29 & 42 & 0.81 & 12 & 15.75 \\
 \hline
 SIM & 1.21 & 9 & 0.94 & 9 & 0.33 & 7 & 2.48 \\
 \hline
 CC & 4.67 & 50 & 4.77 & 52 & 1.92 & 25 & 11.36 \\
 \hline
\end{tabular}
\end{small}
\end{adjustbox}
\caption{Real-world closed-loop validation performance.}
\label{table:real-world-val-eval}
\end{table}

\begin{table}[t!]
\centering
% \begin{adjustbox}{width=\columnwidth,center}
\begin{small}
\begin{tabular}{|c|c c c c c|}
 \hline
 \multirow{2}{*}{Method} & \multicolumn{2}{c}{Scenario 1} & \multicolumn{2}{c}{Scenario 2} & Total \\
 & Tiles & Rules & Tiles & Rules & Tiles \\
 \hline
 HYBRID & \textbf{11} & \cmark & \textbf{19} & \cmark & \textbf{30} \\
 \hline
 REAL & 8 & \cmark & 1 & \xmark & 9 \\
 \hline
 CC & 8 & \cmark & 2 & \xmark & 10 \\
 \hline
\end{tabular}
\end{small}
% \end{adjustbox}
\caption{Real-world closed-loop test performance.}
\label{table:real-world-test-eval}
\end{table}

We also want to emphasize the imperfection of the MSE metric as an indicator for the actual performance, i.e. how well the car drives in an environment, which is usually the end goal. An offline evaluation of control policies is an open research question \cite{Codevilla2018OnOE, Yang2018RealtoVirtualDU}. We provide the MSE results for completeness, but focus our attention on the differences in the end-goal performance.

\subsection{Simulation and Real-world Control Evaluation}

For the end-goal evaluation the goal for the car is to follow the lane either in a simulated or a real-world environment for as many section of the road (tiles) as it can without committing an infraction such as driving off the road or colliding with an object. Each tile is approximately 60 cm long. We have evaluated the HYBRID, REAL, SIM and CC methods.

To validate the simulation performance, we use the Duckietown simulator and pick a map which was not used during the training process. We run each method from 3 starting points (scenarios) and measure the number of tiles driven by the car under a time limit of 15 seconds, or until the first major infraction (driving off the road, e.g.) happens. Survival time marks the end of each run. The results are presented in Table~\ref{table:simulation-val-eval}. All methods drove for 15 seconds without major infractions, and the method trained on SIM method that was trained specifically on the simulation data only drove just 1.8 tiles more than our hybrid approach.

To validate the real-world performance, we use the Duckietown robotarium provided for the use by the organizers. The robotarium represents a complex map with consecutive sharp turns and has diverse background. The robotarium is configured to measure the number of tiles until the first major infraction or the 60 second time limit, and performs the evaluation from 3 different starting points (scenarios). We show the results of this evaluation in Table~\ref{table:real-world-val-eval}. This environment appeared to be more sophisticated than the simulation counterpart. None of the methods reached 60 second mark for any of the starting points which may be a matter of balancing speed and survivability. Comparing the number of tiles, we see that our hybrid approach drove about 3.5 tiles more than the following in the rankings model trained on real-world data only. Due to a poor performance of the SIM approach we remove it from the set of candidates for the real-world closed-loop test evaluation, and test HYBRID, REAL and CC methods only.

Finally, for the real-world closed-loop test evaluation, we report the results of the procedure employed by the Duckietown organizers for the final competition evaluation. It was held in a novel environment, unknown to the participants. We report the number of tiles driven by the car controlled by each method under the 30 second time limit or until the first major infraction. Table~ \ref{table:real-world-test-eval} shows the results for two different starting points (scenarios). As we can see from these results, only the hybrid method was able to complete both scenarios with no infractions. It also demonstrated the highest speed compared to real-world only and classic control methods. In the end, this method took the first place in the competition.

\vspace{-0.09cm}

\section{Conclusion}

In this paper, we present our winning solution to the AI Driving Olympics Lane Following challenge. It follows the imitation learning approach and consists out of a convolutional neural network which is trained on a dataset compiled from data from different sources, such as simulation model and real-world Duckietown vehicle driven by a PD controller, tuned to various conditions, such as different map configuration and lighting.

% We demonstrated the open-loop and closed-loop performance of the method and compared it to the approaches which use either simulation or real-world data only or follow the classic control algorithm. It outperformed other approaches in the real-world closed-loop evaluation while demonstrating competitive results in open-loop and simulation closed-loop cases.

We believe that our approach of emphasizing neurons independence and monitoring generalization performance can offer more robustness to control models that have to perform in diverse environments. We also believe that described approach of imitation learning on data obtained from several algorithms that are fitted to specific environments may yield a single algorithms that will perform well in general.

% Acknowledgements should only appear in the accepted version.
% \section*{Acknowledgements}

% \textbf{Do not} include acknowledgements in the initial version of
% the paper submitted for blind review.

% If a paper is accepted, the final camera-ready version can (and
% probably should) include acknowledgements. In this case, please
% place such acknowledgements in an unnumbered section at the
% end of the paper. Typically, this will include thanks to reviewers
% who gave useful comments, to colleagues who contributed to the ideas,
% and to funding agencies and corporate sponsors that provided financial
% support.

% In the unusual situation where you want a paper to appear in the
% references without citing it in the main text, use \nocite
% \nocite{langley00}

\bibliography{common}
\bibliographystyle{icml2020}

\end{document}